\documentclass[runningheads]{llncs}
\usepackage{graphicx}

\usepackage[accsupp]{axessibility}  

\usepackage{algorithm}
\usepackage{algorithmic}
\usepackage{array}
\usepackage{multirow}
\usepackage{makecell}
\DeclareGraphicsExtensions{.pdf}

\usepackage{tikz}
\usepackage{pifont}
\usepackage{subcaption}

\usepackage{comment}
\usepackage{amsmath,amssymb} 
\usepackage{color}
\usepackage{cite}

\newcommand{\field}[1]{\mathbb{#1}}
\newcommand{\fR}{\field{R}} 
\newcommand{\tw}{\tilde{w}} 

\begin{document}

\title{Prune Your Model Before Distill It} 

\titlerunning{Prune Your Model Before Distill It}
%
\author{Jinhyuk Park\inst{1}\and
Albert No\inst{1}}
\authorrunning{Park and No}
%
\institute{Hongik University, Seoul 04066, Korea\\
\email{c0292601@g.hongik.ac.kr}, \email{albertno@hongik.ac.kr}\\
}

\maketitle

\begin{abstract}
Knowledge distillation transfers the knowledge from a cumbersome teacher to a small student.
Recent results suggest that the student-friendly teacher is more appropriate to distill since it provides more transferable knowledge.
In this work, we propose the novel framework, ``prune, then distill,'' 
that prunes the model first to make it more transferrable and then distill it to the student.
We provide several exploratory examples where the pruned teacher teaches better than the original unpruned networks.
We further show theoretically that the pruned teacher plays the role of regularizer in distillation, which reduces the generalization error.
Based on this result, we propose a novel neural network compression scheme 
where the student network is formed based on the pruned teacher and then apply the ``prune, then distill'' strategy. 
The code is available at \url{https://github.com/ososos888/prune-then-distill}.


\end{abstract}



\section{Introduction}

Recent progress in neural networks (NN) in various tasks highly depends on its over-parameterization,
such as classification~\cite{huang2019gpipe,zagoruyko2016wide}, language understanding~\cite{floridi2020gpt, devlin2019bert},
and  self-supervised learning~\cite{he2020momentum, chen2020big}.
This leads to extensive computational cost and even causes environmental issues~\cite{patterson2021carbon}.
Therefore, neural network compression techniques have received increasing attention,
such as knowledge distillation~\cite{hinton2015distilling, romero2014fitnets, xu2017training}
and pruning~\cite{lecun1990optimal, han2015deep, frankle2019lottery, li2016pruning}. 

Knowledge distillation (KD)~\cite{hinton2015distilling} is a model compression tool 
that transfers the features from a cumbersome network to a smaller network.
At first glance, a powerful teacher with higher accuracy may show better distillation results;
however, Cho and Hariharan~\cite{cho2019efficacy} showed that the less-trained teacher teaches better 
when the student network does not have enough capability.
Lately, a line of works has proposed distillation schemes that focus on a ``student-friendly'' teacher,
which provides more transferrable knowledge to the student network with limited capacity~\cite{park2021learning, Mirzadeh2020ImprovedKD}. 

On the other hand, network pruning~\cite{lecun1990optimal} is another network compression technique 
that effectively removes networks' weights or neurons while maintaining accuracy.
Since pruning simplifies the neural network, we naturally conjecture 
that the pruned teacher provides student-friendly knowledge that is easier to transfer.
This intuition leads us to our main question: {\it can pruning boost the performance of knowledge distillation?}

To answer this question, we propose a new framework, ``prune, then distill,'' consisting of three steps:
1) train the (teacher) network, 2) prune the (teacher) network, and 3) distill the pruned network to the smaller (student) network.
We examine several simple experiments to verify the proposed idea 
that compares the test accuracy of student networks with and without (unstructured) pruning on the teachers' side.
More precisely, We conduct three experiments: 
1) distill VGG19~\cite{simonyan2014very} to VGG11, 
2) distill VGG19 and ResNet18~\cite{he2016deep} to itself (self distillation),
and 3) distill ResNet18 to VGG16 and MobileNetV2~\cite{sandler2018mobilenetv2}.
In all three cases, we observe that the student learned from the pruned teacher 
generally outperforms the student learned from an unpruned teacher.

We then provide theoretical support to answer why the pruned teacher is better in distillation.
Knowledge distillation can be viewed as a label smoothing regularization (LSR)~\cite{yuan2020revisiting, zhou2021rethinking},
which regularizes training by providing a smoother label. 
We find that a teacher trained with regularization provides a smoother label than the original teacher.
This implies that the distillation with a regularized teacher is equivalent to LSR with smoother labels.
Since pruning can be viewed as a regularized model with a sparsity-inducing regularizer~\cite{lejeune2021flip},
we conclude that the pruned teacher regularizes the distillation process.

Based on the observation that pruned teacher provides a better knowledge in distillation,
we then suggest a novel network compression scheme.
When a cumbersome network is provided, we want to compress the network by applying the ``prune, then distill'' strategy.
However, since the distillation transfers knowledge to a {\it given} student network,
the student network architecture design is required.
The main idea of student network construction is matching the teacher and the student layerwise.
We propose a student network with the same depth but fewer neurons so that the number of weights per layer matches the number of nonzero weights of the pruned network in the corresponding layer.
We evaluate the proposed compression scheme with extensive experiments. 

We summarize our contributions as:
\begin{itemize}
\item We propose a novel framework, ``prune, then distill,'' that prunes teacher networks before distillation.
\item We examine experiments that verify unstructured pruning on the teacher can boost the performance of knowledge distillation.
\item We also provide a theoretical analysis that the distillation from a pruned teacher is effectively a label smoothing regularization with smoother labels.
\item We propose a novel network compression that constructs the student network based on the pruned teacher,
    then apply the ``prune, then distill'' strategy.
\end{itemize}


\section{Related Works}
This section is devoted to prior works on neural network (NN) compression that are related to our work.
In particular, we focus on knowledge distillation and network pruning.
Note that there are other NN compression techniques such as quantization~\cite{li2016ternary, scalableQuant},
coding~\cite{DeepCABAC, havasi2018minimal}, and matrix factorization~\cite{low_rank1, Idelbayev_low_rank3}.

\subsection{Knowledge Distillation}
Knowledge distillation (KD)~\cite{hinton2015distilling} transfers the knowledge from the strong teacher network to a smaller student network.
The student network is trained with soft targets provided by the teacher network and some intermediate features
\cite{romero2014fitnets, zagoruyko2016paying, yang2020distilling}.
There are variations of KD such as KD using GAN~\cite{xu2017training},
Jacobian matching KD~\cite{czarnecki2017sobolev, srinivas2018knowledge},
distillation of activation boundaries~\cite{heo2019knowledge}, contrastive distillation~\cite{tian2019contrastive},
and distillation from graph neural networks~\cite{yang2020distilling, jing2021amalgamating}.

Recently, many works have reported that the large gap between student and teacher 
causes degradation in student network performance~\cite{Mirzadeh2020ImprovedKD}.
Cho and Hariharan showed that the less-trained network transfers better knowledge to a small network~\cite{cho2019efficacy}.
Park et al.~\cite{park2021learning} proposed a student-aware teacher learning to transfer the teacher's knowledge effectively.
In this paper, we provide an extremely simple way to generate a student-friendly teacher using unstructured pruning.

\subsection{Pruning}
There are two main branches of pruning: 1) unstructured pruning, which prunes individual weights,
and 2) structured pruning, which prunes neurons (in most cases, channels of convolutional neural networks). 
Although both approaches share a similar idea, these two strategies have been developed independently.

\noindent{\bf Unstructured pruning: }
Unstructured pruning~\cite{lecun1990optimal} removes NN components in weight-level while maintaining the number of neurons in the network.
A general pruning pipeline consists of three steps:
1) train a large network, 2) prune weights (or neurons) based on its own rule, then 3) fine-tune the pruned model.
The iterative magnitude pruning (IMP) technique, which iteratively applies magnitude-based pruning and fine-tuning, shows remarkable performance~\cite{han2015deep}.
Lottery ticket rewinding (LTR), an iterative magnitude pruning method with weight rewinding,
is highly successful~\cite{frankle2019lottery, frankle2020linear}.
Recently, IMP with learning rate (LR) rewinding, which repeats the learning rate schedule,
shows better results in bigger networks~\cite{renda2020comparing}.
However, the network architecture after unstructured pruning remains the same (i.e., number of channels per layer).
It is hard to fully enjoy the benefit of a pruned network without dedicated hardware~\cite{han2016eie}.

\noindent{\bf Structured pruning: }
Structured pruning removes NN parameters at the level of neurons (mostly channels)
\cite{li2016pruning,anwar2017structured,luo2017thinet,liu2019metapruning,su2021bcnet}.
It provides a smaller network with efficient network architecture,
and we can save computational resources without designing dedicated hardware or libraries.
Like magnitude-based unstructured pruning, the most naive method is to prune filters based on weights~\cite{li2016pruning, he2018soft}.
Another approach is adding an extra regularizer that induces sparsity while training~\cite{wen2016learning, zhou2016less, he2017channel}.
Liu et al.~\cite{liu2017learning} and Ye et al.~\cite{ye2018rethinking} proposed 
the structured pruning scheme based on batch normalization (BN) scale factor of filters.
Zhuang et al.~\cite{zhuang2020neuron} adds polarization regularizer to structured pruning with BN scale factor.
However, due to the structural constraint, the pruned network has more weights (parameters) than unstructured pruning~\cite{liu2018rethinking}.


\section{Prune, then Distill}\label{sec:pruneKD}
\begin{figure}[t]
\centering
\centering
\includegraphics[width=.99\textwidth]{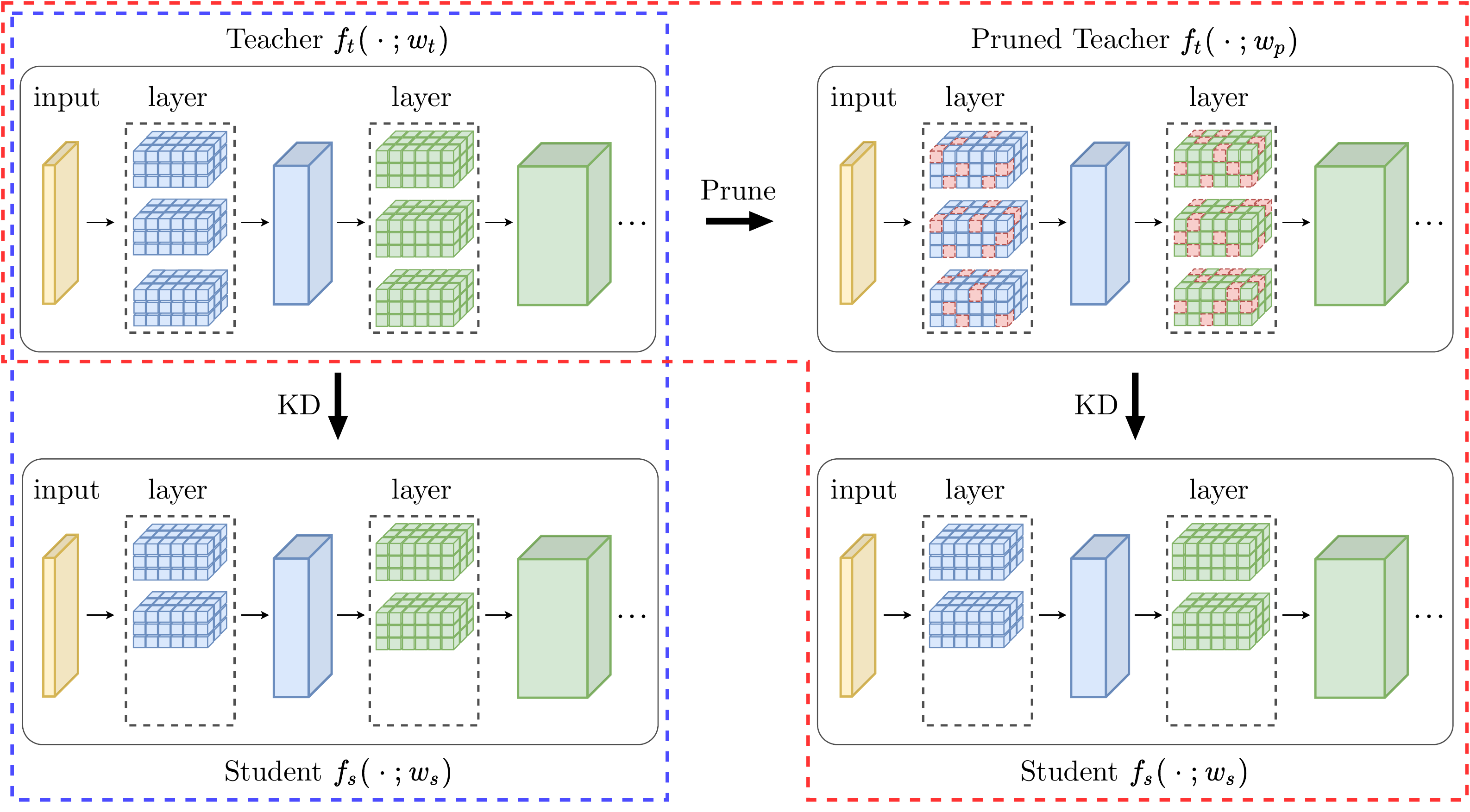}
\caption{Overview of the ``prune, then distill'' strategy.
Instead of distilling directly from the teacher to the student (blue dotted box),
we prune the teacher first, then distill from the pruned teacher to the student (red dotted box).
}
\label{fig:KD}
\end{figure}

\subsection{Exploratory Experiments}
We conduct experiments to verify the effectiveness of pruned teachers in KD.
Instead of distilling the teacher network directly (dotted-blue block in Figure~\ref{fig:KD}),
we first (unstructured) prune the teacher network and then distill to the student network (dotted-red block in Figure~\ref{fig:KD}).

\noindent{\bf Setups: }
We mainly considers VGG~\cite{simonyan2014very} and ResNet~\cite{he2016deep} for the teacher network,
where VGG is trained on the CIFAR100 dataset~\cite{krizhevsky2009learning}
and ResNet is trained on the TinyImageNet dataset~\cite{le2015tiny}.
The TinyImageNet dataset is a subset of resized ($3\times64\times64$) ImageNet dataset~\cite{deng2009imagenet}.
We reserve 10\% of the data as a validation set in all training.
We apply unstructured pruning that removes more weights, more precisely LR rewinding~\cite{renda2020comparing}, to prune the teacher model.
In LR rewinding, we set the ratio of epochs by 0.65 for VGG-CIFAR100.
In other words, we train the VGG19 for 200 epochs initially,
then rewind the learning rates and retrains (fine-tuning) the network for 130 epochs (65\%).
Note that the different ratios from 0.6 to 0.9 do not make significant differences in pruning, and we use the ratio of 0.5 for ResNet-TinyImageNet. 
For a fair comparison, we train (and distill) networks with enough epochs and halt the training at their best performance on validation dataset.
All test accuracies are the average of three independent experiments, and we also provide the standard deviation.

For simplicity, we use the vanilla KD~\cite{hinton2015distilling} with appropriate balancing parameter $\alpha$ and temperature $\tau$.
The balancing parameter $\alpha$ represents the ratio of two objectives (distill loss and hard-target loss).
The temperature $\tau$ is a softening parameter, where higher $\tau$ produces a softer target.
In the experiment, we fix the parameters by $\alpha = 0.95$ and $\tau = 10$.
More detailed training parameters are provided in Appendix.
Note that the purpose of experiments is not achieving the best test accuracy
but to compare between {\it distilling from a pruned network} and {\it distilling from an unpruned network}.
Thus, hyperparameters, as well as network architectures, are not optimized for test accuracies.
Instead, we use as-is settings for a {\it fair} comparison between the pruned teacher and the unpruned teacher.
For example, we follow default settings for MobileNetV2~\cite{sandler2018mobilenetv2} and ResNet18~\cite{he2016deep} 
optimized for ImageNet dataset~\cite{deng2009imagenet}, while we use TinyImageNet dataset~\cite{le2015tiny}.

\noindent{\bf Distill VGG19 to VGG11: }
We set VGG19~\cite{simonyan2014very} as a teacher network and VGG11 as a student network.
The network architecture of VGG is unchanged except for the number of fully connected (FC) layers,
where our VGG has a single FC layer (which is commonly used for CIFAR10 data).
Then, we compare the KD results on the CIFAR100 dataset~\cite{krizhevsky2009learning}
between the regular VGG19 teacher and the pruned VGG19 teacher.
We prune the teacher network with three sparsity levels: 36\% sparsity (36\% of weights are removed),
59\% sparsity, and 79\% sparsity.

\begin{table*}[t]
\begin{center}
\caption{
Knowledge distillation from VGG19 to VGG11 on CIFAR100 with teacher pruning.
VGG19DBL is the VGG19 with $2\times$ more filters per layer.
Teacher ``None'' indicates the student is trained without a teacher,
while the pruning ratio ``None'' means the distillation from the unpruned teacher.
}
\label{tab:vgg19-vgg11}
\newcolumntype{A}{>{\centering}p{0.20\textwidth}}
\newcolumntype{B}{>{\centering}p{0.17\textwidth}}
\newcolumntype{C}{>{\centering}p{0.17\textwidth}}
\newcolumntype{D}{>{\centering}p{0.18\textwidth}}
\newcolumntype{E}{>{\centering\arraybackslash}p{0.18\textwidth}}
\begin{tabular}{ABC|DE}
\Xhline{1pt}
Teacher & \makecell{ Pruning \\Ratio} & \makecell{Teacher\\ Accuracy } & Student & \makecell{Student\\Accuracy}\\\Xhline{1pt}
None          & -            & - & VGG11         & 69.51 $\pm$ 0.24\\\hline
\multirow{4}{*}{VGG19}  & None          & 73.13 & VGG11     & 72.02 $\pm$ 0.27\\
           & 36\%       &  73.30  & VGG11     & 72.76 $\pm$ 0.10\\
           & 59\%       &   72.25 & VGG11     & 72.59 $\pm$ 0.32\\
           & 79\%       &  73.43  & VGG11     & 72.67 $\pm$ 0.34\\\hline
\multirow{4}{*}{VGG19DBL}  & None      &  74.44 & VGG11   & 71.81 $\pm$ 0.29\\
  & 36\%         & 73.46 & VGG11    & 72.01 $\pm$ 0.11\\
  & 59\%         & 73.24 & VGG11    & 72.40 $\pm$ 0.25\\
  & 79\%         & 73.50 & VGG11    & 72.48 $\pm$ 0.19\\
\Xhline{1pt}
\end{tabular}
\end{center}
\end{table*}

Surprisingly, as shown in Table~\ref{tab:vgg19-vgg11}, VGG11 with pruned VGG19 consistently outperforms the one with the unpruned teacher.
Table~\ref{tab:vgg19-vgg11} also provides results when the teacher network is VGG19DBL, 
with $2\times$ many channels in each layer.
In both cases, the pruned teacher shows better performance.

\begin{table*}[t]
\begin{center}
\caption{Self distillation of VGG19 and ResNet18 with teacher pruning. 
DBL model has the same model structure with $2\times$ more filters per layer.
Teacher ``None'' indicates the student is trained without a teacher,
while the pruning ratio ``None'' means the distillation from the unpruned teacher.
}
\label{tab:self-distillation}
\newcolumntype{A}{>{\centering}p{0.20\textwidth}}
\newcolumntype{B}{>{\centering}p{0.16\textwidth}}
\newcolumntype{C}{>{\centering}p{0.16\textwidth}}
\newcolumntype{D}{>{\centering}p{0.20\textwidth}}
\newcolumntype{E}{>{\centering\arraybackslash}p{0.18\textwidth}}
\begin{tabular}{ABC|DE}
\Xhline{1pt}
Teacher & \makecell{ Pruning \\Ratio} & \makecell{Teacher\\ Accuracy } & Student & \makecell{Student\\Accuracy}\\\Xhline{1pt}

None & - & - & VGG19    & 72.76 $\pm$ 0.33\\\hline
\multirow{4}{*}{VGG19}  & None         & 73.13 & VGG19     & 73.74 $\pm$ 0.20\\
           & 36\%         & 73.30 & VGG19     & 74.10 $\pm$ 0.26\\
           & 59\%         & 72.25 & VGG19     & 74.26 $\pm$ 0.37\\
           & 79\%         & 73.43 & VGG19     & 74.35 $\pm$ 0.10\\\hline

None & - & - & VGG19DBL     & 74.62 $\pm$ 0.21\\\hline
\multirow{4}{*}{VGG19DBL}  & None         & 74.44 & VGG19DBL     & 74.78 $\pm$ 0.37\\
  & 36\%       & 73.46    & VGG19DBL     & 75.16 $\pm$ 0.44\\
  & 59\%        &  73.24 & VGG19DBL     & 75.26 $\pm$ 0.77\\
  & 79\%       & 73.50    & VGG19DBL     & 75.05 $\pm$ 0.92\\\hline
  
None & - & - & ResNet18    &  57.75 $\pm$ 0.24 \\\hline
\multirow{4}{*}{ResNet18}  & None      & 57.75     & ResNet18     & 57.97 $\pm$ 0.10\\
  & 36\%       & 57.66     & ResNet18     &  59.39 $\pm$ 0.21 \\
  & 59\%       & 57.58    & ResNet18     &  58.99 $\pm$ 0.26 \\
  & 79\%       & 57.32    & ResNet18     &  59.33 $\pm$ 0.18 \\\hline

None & - & - & ResNet18DBL    & 60.21 $\pm$ 0.24\\\hline
\multirow{4}{*}{ResNet18DBL}  & None     &  60.46    & ResNet18DBL     & 61.35 $\pm$ 0.02\\
  & 36\%       & 61.97    & ResNet18DBL     & 63.03 $\pm$ 0.38\\
  & 59\%       & 61.80    & ResNet18DBL     & 63.19 $\pm$ 0.21\\
  & 79\%        &  61.66 & ResNet18DBL     & 63.16 $\pm$ 0.02\\

\Xhline{1pt}
\end{tabular}
\end{center}
\end{table*}

\noindent{\bf Self distillation: }
Motivated by~\cite{NEURIPS2020_f3ada80d, yuan2020revisiting}, we conduct the self distillation experiment,
where the teacher and the student share the same model.
We consider VGG19 and VGG19DBL with CIFAR100 dataset, where ResNet18 and ResNet18DBL are trained on the TinyImageNet dataset.
Table~\ref{tab:self-distillation} shows the test accuracies of
1) the model without KD, 2) the model learned from the unpruned teacher, and 3) the model learned from the pruned teacher.
Similar to other experiments, we also observe the consistent result where the pruned model teaches better than the unpruned teacher.
Note that learning from unpruned network also increases the test accuracy (compared to the one without a teacher);
however, the gain with the pruned teacher is more significant.

\begin{table*}[t]
\begin{center}
\caption{Distillation from ResNet18 to MobileNetV2 and VGG16 with teacher pruning.
Teacher ``None'' indicates the student is trained without a teacher,
while the pruning ratio ``None'' means the distillation from the unpruned teacher.
}
\label{tab:cross-platform}
\newcolumntype{A}{>{\centering}p{0.20\textwidth}}
\newcolumntype{B}{>{\centering}p{0.16\textwidth}}
\newcolumntype{C}{>{\centering}p{0.16\textwidth}}
\newcolumntype{D}{>{\centering}p{0.20\textwidth}}
\newcolumntype{E}{>{\centering\arraybackslash}p{0.18\textwidth}}
\begin{tabular}{ABC|DE}
\Xhline{1pt}
Teacher & \makecell{ Pruning \\Ratio} & \makecell{Teacher\\ Accuracy } & Student & \makecell{Student\\Accuracy}\\\Xhline{1pt}
None & -          & - & VGG16      & 53.31 $\pm$ 0.45\\\hline
\multirow{4}{*}{ResNet18}  & None        & 57.75  & VGG16   & 54.75 $\pm$ 0.29\\
           & 36\%         & 57.66  & VGG16   & 56.35 $\pm$ 0.35\\
           & 59\%         & 57.58 & VGG16    & 55.86 $\pm$ 0.04\\
           & 79\%        &  57.32 & VGG16    & 56.49 $\pm$ 0.15\\\hline

None & -          & - & MobileNetV2     & 50.79 $\pm$ 0.44\\\hline
\multirow{4}{*}{ResNet18}  & None        & 57.75  & MobileNetV2   & 56.10 $\pm$ 0.23\\
  & 36\%        &  57.66 & MobileNetV2   & 56.73 $\pm$ 0.24\\
  & 59\%       & 57.58    & MobileNetV2   & 56.73 $\pm$ 0.43\\
  & 79\%        &  57.32 & MobileNetV2   & 57.20 $\pm$ 0.25\\
\Xhline{1pt}
\end{tabular}
\end{center}
\end{table*}

\noindent{\bf Distill ResNet18 to VGG and MobileNet: }
We also investigate the KD from the pruned teacher when the student and the teacher have different network architectures.
Specifically, we consider the TinyImageNet dataset, where the teacher is ResNet18
and students are VGG16 and MobileNetV2~\cite{sandler2018mobilenetv2}.
Table~\ref{tab:cross-platform} compares the test accuracies of 
1) student without a teacher, 2) student learned from the unpruned teacher, 
and 3) student learned from the pruned teacher.
Consistently, we observe the better KD performance when the teacher is pruned.
This implies that the better distillation is not limited to the case of the similar architecture between teacher and student networks.

\noindent{\bf Remark: }
One might suspect that better distillation result is due to higher accuracy of the teacher,
where the pruned model often achieves better accuracy~\cite{frankle2019lottery}.
However, the higher accuracy of the teacher network does not guarantee better results in distillation~\cite{stanton2021does}.
Also, the pruned teacher works better even when test accuracy is lower than the unpruned teacher.
For example, pruning decreases the test accuracy of the teacher network in ResNet18-TinyImageNet,
where we observe that the pruned teacher transfers the knowledge better.
This implies that the pruned teacher is better not because it has higher accuracy, but it provides better transferable knowledge.

We also investigate the agreement between the teacher and the student's prediction (details provided in Appendix).
As shown by Stanton et al.~\cite{stanton2021does}, we observe that the agreement and the accuracy behave independently.
For example, in VGG19 self distillation experiments, the pruned teacher provides a higher agreement, and the corresponding student has a higher accuracy;
however, in ResNet18 self distillation, the pruned teacher shows lower agreement although the student's accuracy is higher. 
It implies that some students mimic the teacher better but perform worse.
This result supports our theory that distillation indirectly helps the training student models with additional regularization.

\subsection{Pruned Teacher as a Regularizer}
In this section, we provide a theoretical analysis on the pruned teacher in KD.
We first point out that the teacher trained with a regularizer provides an additional regularization during distillation.

Let $\{(x_i, y_i)\}_{i=1}^N$ be the dataset where the label $y_i$ takes value from the set $\{1, 2, \dots, K\}$.
We are interested in a classification model which outputs a $K$-dimensional probability distributions.
Let $f_{true}(x_i)\in \fR^K$ be the one-hot encoded vector 
where $f_{true}(x_i)[y_i]=1$ for the ground-truth label $y_i$ and $f_{true}(x_i)[y']=0$ for all $y'\neq y_i$.
We further let $f_t(x; w)$ be the output of the teacher network when the input is $x$ and the weight is $w$.
Then, we train the teacher $f_t(\cdot; w)$ and achieve $w_t$ that minimizes the cross entropy loss
\begin{align}
    L_{CE}(w) = \frac{1}{N}\sum_{i=1}^N H(f_{true}(x_i), f_t(x_i;w)), \label{eq:ce loss}
\end{align}
where the cross-entropy loss is defined by $H(p_1, p_2) = -\sum_{k=1}^K p_1[k]\log p_2[k]$.

Similarly, $f_s(x;\tw)$ is the output of the student network when the input is $x$ and the weight is $\tw$.
For the temperature $\tau=1$, the knowledge distillation loss is given by
\begin{align}
    L_{KD}(\tw) = \frac{1}{N}\sum_{i=1}^N (1-\alpha) H(f_{true}(x_i) , f_s(x_i;\tw)) + \alpha H(f_t(x;w_t), f_s(x;\tw)).
\end{align}
Yuan et al.~\cite{yuan2020revisiting} showed that the KD is equivalent to label smoothing regularization (LSR).
More precisely, the author showed that
\begin{align}
    L_{KD}(\tw) = \frac{1}{N} \sum_{i=1}^N H(f^{(\alpha)}_m(x_i;w_t), f_s(x_i;\tw)),
\end{align}
where $f^{(\alpha)}_m(x;w_t) = (1-\alpha) f_{true}(x) + \alpha f_t(x; w_t)$,
and therefore KD is equivalent to label smoothing regularization with smoothed label distribution $f^{(\alpha)}_m(x;w_t)$.

We then consider the case where the teacher is trained with a regularizer $R(w)$.
The regularized teacher $f_t(\cdot;w_p)$ is obtained by minimizing
\begin{align}
    L_{REG}(w) = \frac{1}{N}\sum_{i=1}^N H(f_{true}(x_i), f_t(x_i;w)) + R(w), \label{eq:re loss}
\end{align}
i.e., $L_{REG}(w_p) = \min_w L_{REG}(w)$.
Since $L_{CE}(w_t) = \min_w L_{CE}(w)$, we have
\begin{align}
    \frac{1}{N}\sum_{i=1}^N H(f_{true}(x_i), f_t(x_i;w_t)) \leq& \frac{1}{N}\sum_{i=1}^N H(f_{true}(x_i), f_t(x_i;w_p))\\
    \frac{1}{N}\sum_{i=1}^N H(f_{true}(x_i), f_t(x_i;w_p)) + R(w_p) \leq& \frac{1}{N}\sum_{i=1}^N H(f_{true}(x_i), f_t(x_i;w_t)) + R(w_t)
\end{align}
which implies
\begin{align}
    0 \leq \frac{1}{N}\sum_{i=1}^N \log\frac{f_t(x_i;w_t)[y_i]}{f_t(x_i;w_p)[y_i]}
    \leq R(w_t) - R(w_p) \label{eq:regularized teacher}
\end{align}
Thus, $f_t(x_i;w_t)[y_i]$ is larger than $f_t(x_i;w_p)[y_i]$ on average.

Recall that the distillation from $f_t(x_i;w_t)$ is equivalent to label smoothing regularization with smoothed label distribution
$f^{(\alpha)}_m(x;w_t) = (1-\alpha) f_{true}(x) + \alpha f_t(x, w_t)$.
If we distill from $f_t(x_i; w_p)$ to the student,
then it is essentially label smoothing regularization with a new smoothed label distribution
$f^{(\alpha)}_{m}(x;w_p) = (1-\alpha) f_{true}(x) + \alpha f_t(x, w_p)$.
Since Eq.~\eqref{eq:regularized teacher} implies that the new smoothed distribution $f^{(\alpha)}_{m}(x_i;w_p)$ 
has a smaller weight at the true label $y_i$ on average,
we can conclude that $f^{(\alpha)}_{m}(x_i;w_p)$ is {\it smoother}\footnote{
Instead of label's self-entropy, we measure the smoothness with true label's weight.}
than $f^{(\alpha)}_m(x_i;w_t)$.
In other words, the regularization in teacher training also regularizes student distillation further.
Note that Eq.~\eqref{eq:regularized teacher} provides an upper bound of the ratio 
between the teacher's output and the regularized teacher's output at the true label.
This effectively measures the smoothness of a smoothed label in label smoothing regularization.

The pruning can be viewed as a solution of the empirical risk minimization problem with sparsity-inducing regularization~\cite{lejeune2021flip}.
Thus, the distillation from the pruned teacher is a label smoothing regularization with smoother label distribution,
which reduces a generalization error.

\section{Transferring Knowledge of Sparsity}

Based on the observation that the pruned teacher transfers the better knowledge,
we propose a novel network compression framework that learns from the (unstructured) pruned network.
The critical challenge is a student network architecture design to learn effectively from the pruned teacher.

More formally, let $f_t(\cdot;w_t)$ be a cumbersome network to compress, and the goal is to compress it to a smaller network $f_s(\cdot;w_s)$.
In the previous section, we considered the distillation to a given student network.
On the other hand, in this section, we provide a detailed architecture design for a student network $f_s$ based on the pruned teacher $f_t(\cdot;w_p)$.

\begin{figure}[t]
\centering
\includegraphics[width=.99\textwidth]{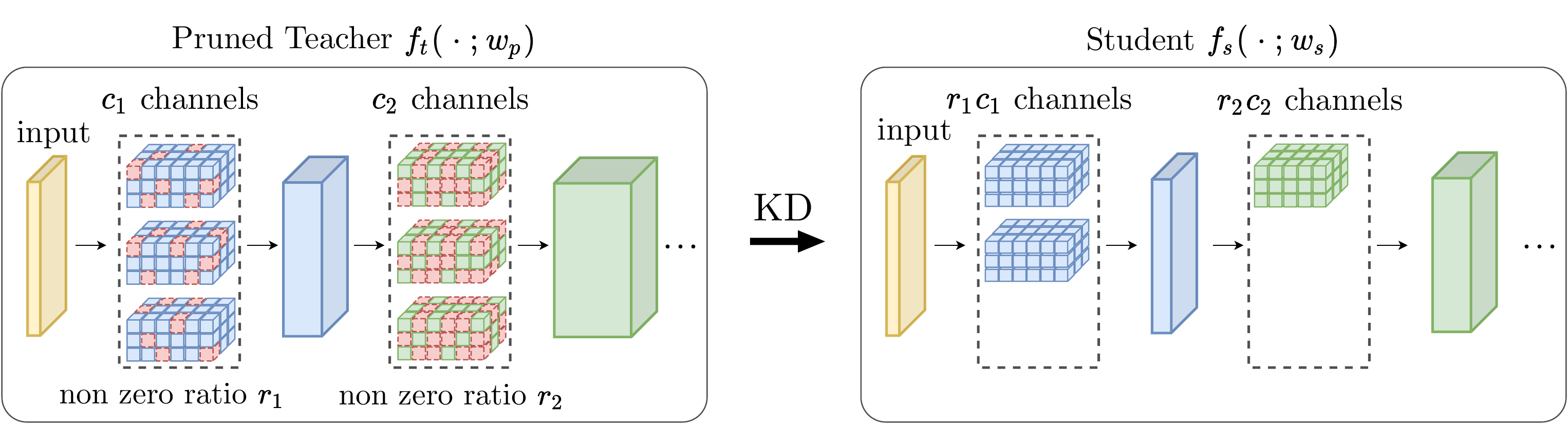}
\caption{Student network design. 
The number of channels of the student network is adjusted so that each layer's parameters match the number of nonzero parameters in each layer of the pruned teacher.}
\label{fig:Student network design}
\end{figure}

On top of the ``prune, then distill'' as described in Figure~\ref{fig:KD},
we add student network architecture design.
The key idea of student network design is that the pruned teacher can also provide sparsity knowledge.
We construct the narrower student where each layer matches the corresponding layer of the pruned teacher.
More precisely, the student network has the same depth,
but the number of channels per layer is reduced so that the number of weights is (approximately) equal 
to the number of remaining parameters in the pruned teacher (as described in Figure~\ref{fig:Student network design}).
The intuition is to build a student network where each layer has enough capacity to learn from the pruned teacher.
The rigorous construction of the student network is described in Appendix.
Thus, the proposed compression algorithm has four steps:
\begin{enumerate}
\item Train the original network and obtain $f_t(\cdot;w_t)$.
\item Apply the unstructured pruning and obtain pruned network $f_t(\cdot;w_p)$.
\item Construct $f_s$ based on each layer's sparsity of the pruned network $f_t(\cdot;w_p)$.
\item Distill the pruned network $f_t(\cdot;w_p)$ to the student $f_s(\cdot;w_s)$.
\end{enumerate}

Note that the above framework does not depend on the specific choice of distillation or pruning method.
In Section~\ref{sec:Experiments}, we apply LR rewinding~\cite{renda2020comparing} to prune the model,
and apply the vanilla KD~\cite{hinton2015distilling} to distill the pruned teacher.

The proposed scheme transfers knowledge from the sparse network (from unstructured pruning)
to a network with fewer channels to reduce the number of channels further.
This is similar to residual distillation~\cite{li2020residual} which removes unwanted parts (residual connections) of residual networks.
In our setting, we remove unwanted parts (more channels) of unstructured pruning by merging sparse filters into fewer filters via KD.

Note that our compression framework can be viewed as structured pruning since it effectively removes neurons (channels) of a given network.
Since structured pruning is nearly an architecture search algorithm~\cite{liu2018rethinking},
the proposed framework suggests a novel network architecture search algorithm that learns from unstructured pruning.
Recall that recent global unstructured pruning algorithms~\cite{lee2021layeradaptive} 
(where the pruning scheme actively determines the pruning ratio for each layer) outperform precisely designed layerwise sparsity selection schemes.

\section{Experiments}\label{sec:Experiments}
In this section, we present our experimental results verifying the proposed algorithm.
Similar to Section~\ref{sec:pruneKD}, we compare test accuracies of three scenarios:
1) train student network without a teacher,
2) distill the pruned teacher to the student network,
and 3) distill the original (unpruned) teacher to the student network.
To maintain the consistency of experiments, we use the same training, pruning, and distillation procedure 
and the same network hyperparameters for all three scenarios (mostly from Section~\ref{sec:pruneKD}).
All test accuracies are the average of three independent experiments, and we also provide the standard deviation.

\subsection{Results}
For the VGG-CIFAR100 experiment, we use VGG19 with batch normalization as a teacher.
In the proposed framework, we apply LR rewinding to obtain the pruned VGG19s with target sparsity 36\%, 59\%, and 79\%.
The test accuracy of the pruned teacher 
is similar to the baseline model (VGG19) or slightly higher.
We construct the student network as described in the previous section.
Let VGG19-ST36, VGG19-ST59, and VGG19-ST79 denote the student networks with fewer channels 
that correspond to pruned teachers with pruning ratios 36\%, 59\%, and 79\%, respectively.
We also run the same experiment with VGG19DBL (with $2\times$ more channels per layer).
Similar to VGG19, let VGG19DBL-ST36, VGG19DBL-ST59, and VGG19DBL-ST79 denote student networks 
that correspond to pruned teachers with pruning ratios 36\%, 59\%, and 79\%, respectively.

\begin{table*}[t]
\caption{Performance of the proposed compression algorithm on VGG19 with CIFAR100.
VGG19-ST(X) is the constructed student network based on the proposed algorithm from X\% pruned teacher.
Teacher ``None'' indicates the student is trained without a teacher,
while the pruning ratio ``None'' means the distillation from the unpruned teacher.
}
\label{tab:algorithm-vgg}
\begin{center}
\newcolumntype{A}{>{\centering}p{0.20\textwidth}}
\newcolumntype{B}{>{\centering}p{0.14\textwidth}}
\newcolumntype{C}{>{\centering}p{0.14\textwidth}}
\newcolumntype{D}{>{\centering}p{0.24\textwidth}}
\newcolumntype{E}{>{\centering\arraybackslash}p{0.18\textwidth}}
\begin{tabular}{ABC|DE}
\Xhline{1pt}
Teacher & \makecell{ Pruning \\Ratio} & \makecell{Teacher\\ Accuracy } & Student & \makecell{Student\\Accuracy}\\\Xhline{1pt}

None & -          & - & VGG19-ST36         & 72.32 $\pm$ 0.12\\\hline
\multirow{2}{*}{VGG19}  & None        & 73.13  & VGG19-ST36     & 73.52 $\pm$ 0.20\\
           & 36\%        & 73.30  & VGG19-ST36     & 73.77 $\pm$ 0.16\\\hline
   
None & -          & - & VGG19-ST59         & 71.80 $\pm$ 0.18\\\hline
\multirow{2}{*}{VGG19}  & None       &  73.13 & VGG19-ST59     & 73.18 $\pm$ 0.10\\
           & 59\%         & 72.25 & VGG19-ST59     & 73.81 $\pm$ 0.10\\\hline

None & -          & - & VGG19-ST79         & 70.89 $\pm$ 0.14\\\hline
\multirow{2}{*}{VGG19}  & None       &  73.13  & VGG19-ST79     & 72.42 $\pm$ 0.16\\
           & 79\%       &  73.43  & VGG19-ST79     & 73.39 $\pm$ 0.11\\\hline
\Xhline{1pt}

None & -          & - & VGG19DBL-ST36         & 74.39 $\pm$ 0.02\\\hline
\multirow{2}{*}{VGG19DBL}  & None      &  74.44   & VGG19DBL-ST36     & 74.62 $\pm$ 0.34\\
           & 36\%       &  73.46  & VGG19DBL-ST36     & 75.40 $\pm$ 0.18\\\hline

None & -          & - & VGG19DBL-ST59         & 74.06 $\pm$ 0.22\\\hline
\multirow{2}{*}{VGG19DBL}  & None     &   74.44   & VGG19DBL-ST59     & 74.67 $\pm$ 0.24\\
           & 59\%      &   73.24  & VGG19DBL-ST59     & 75.09 $\pm$ 0.23\\\hline

None & -          & - & VGG19DBL-ST79         & 73.81 $\pm$ 0.45\\\hline
\multirow{2}{*}{VGG19DBL}  & None    &  74.44     & VGG19DBL-ST79     & 74.16 $\pm$ 0.04\\
           & 79\%     &   73.50   & VGG19DBL-ST79     & 75.19 $\pm$ 0.31\\\hline

\Xhline{1pt}
\end{tabular}
\end{center}
\end{table*}

\begin{table*}[t]
\caption{Performance of the proposed compression algorithm on ResNet18 with TinyImageNet.
ResNet18-ST(X) is the constructed student network based on the proposed algorithm from X\% pruned teacher.
Teacher ``None'' indicates the student is trained without a teacher,
while the pruning ratio ``None'' means the distillation from the unpruned teacher.
}
\label{tab:algorithm-resnet}
\begin{center}
\newcolumntype{A}{>{\centering}p{0.20\textwidth}}
\newcolumntype{B}{>{\centering}p{0.12\textwidth}}
\newcolumntype{C}{>{\centering}p{0.12\textwidth}}
\newcolumntype{D}{>{\centering}p{0.28\textwidth}}
\newcolumntype{E}{>{\centering\arraybackslash}p{0.18\textwidth}}
\begin{tabular}{ABC|DE}
\Xhline{1pt}
Teacher & \makecell{ Pruning \\Ratio} & \makecell{Teacher\\ Accuracy } & Student & \makecell{Student\\Accuracy}\\\Xhline{1pt}
None & - & - & ResNet18-ST36 & 56.44 $\pm$ 0.26\\\hline
\multirow{2}{*}{ResNet18}  & None           & 57.75 & ResNet18-ST36    & 57.74 $\pm$ 0.22\\
           & 36\%          & 57.66 & ResNet18-ST36  & 58.75 $\pm$ 0.19\\\hline

None & - & - & ResNet18-ST59    & 55.93 $\pm$ 0.32\\\hline
\multirow{2}{*}{ResNet18}  & None          & 57.75 & ResNet18-ST59  & 56.70 $\pm$ 0.35\\
           & 59\%          & 57.58 & ResNet18-ST59  & 57.76 $\pm$ 0.31\\\hline

None & - & - & ResNet18-ST79    & 54.48 $\pm$ 0.53\\\hline
\multirow{2}{*}{ResNet18}  & None         & 57.75 & ResNet18-ST79    & 55.65 $\pm$ 0.24\\
           & 79\%          & 57.32 & ResNet18-ST79    & 56.23 $\pm$ 0.16\\\hline

\Xhline{1pt}

None & - & - & ResNet18DBL-ST36     & 59.88 $\pm$ 0.30 \\\hline
\multirow{2}{*}{ResNet18DBL}  & None        & 60.46  & ResNet18DBL-ST36    & 61.02 $\pm$ 0.15 \\
           & 36\%         & 61.97 & ResNet18DBL-ST36    &  62.33 $\pm$ 0.21\\\hline

None & - & - & ResNet18DBL-ST59     & 58.81 $\pm$ 0.28 \\\hline
\multirow{2}{*}{ResNet18DBL}  & None         & 60.46 & ResNet18DBL-ST59    & 60.99 $\pm$ 0.27\\
           & 59\%        &  61.80 & ResNet18DBL-ST59     &  62.41 $\pm$ 0.52\\\hline

None & - & - & ResNet18DBL-ST79      & 57.79 $\pm$ 0.14 \\\hline
\multirow{2}{*}{ResNet18DBL}  & None         & 60.46 & ResNet18DBL-ST79    & 60.60 $\pm$ 0.26\\
           & 79\%        &  61.66 & ResNet18DBL-ST79    & 61.87 $\pm$ 0.27 \\

\Xhline{1pt}
\end{tabular}
\end{center}
\end{table*}

For the ResNet-TinyImageNet experiment, we use ResNet18 as a teacher.
The base ResNet18 is an unpruned teacher model where the test accuracy is 57.75\%.
The pruned ResNet18 is a teacher in the proposed framework where we apply LR rewinding with target sparsity 36\%, 59\%, and 79\%.
Notably, the pruned teacher's test accuracy is lower than the unpruned network, unlike the VGG-CIFAR100 setup.
Similar to VGG-CIFAR100, let ResNet18-ST36, ResNet18-ST59, and ResNet18-ST79 denote the student networks 
that correspond to the pruned teacher with pruning ratios 36\%, 59\%, and 79\%, respectively.

Table~\ref{tab:algorithm-vgg} and Table~\ref{tab:algorithm-resnet} show the test accuracies of the student network.
For comparison, we also provide test accuracies when the same student network is trained without a teacher.
In all settings, the proposed scheme outperforms the student network trained from scratch by huge margin.

\subsection{Ablation Study}

{\bf Learning from the unpruned teacher: }
Table~\ref{tab:algorithm-vgg} and Table~\ref{tab:algorithm-resnet} 
also provide the KD result from the unpruned teacher with the same student networks.
Similar to Section~\ref{sec:pruneKD},
it is consistent that the pruned teacher (with matching sparsity) provide better KD.

\noindent{\bf Alternative student network design: }
For VGG19(DBL) teacher, we manually designed students VGG19-CL1 and VGG19-CL2.
These networks have the same depth, but the number of channels is adjusted,
where the number of network parameters is (approximately) half of the original network.
VGG19-CL1 removes channels uniformly across the layer, and VGG19-CL2 removes channels unevenly.
The detailed network architecture is provided in Appendix.
\begin{table*}[t]
\begin{center}
\caption{Knowledge distillation to manually designed student networks.
VGG19DBL is the VGG19 with $2\times$ more filters per layer.
Teacher ``None'' indicates the student is trained without a teacher,
while the pruning ratio ``None'' means the distillation from the unpruned teacher.
}
\label{tab:custom-layer}
\newcolumntype{A}{>{\centering}p{0.20\textwidth}}
\newcolumntype{B}{>{\centering}p{0.14\textwidth}}
\newcolumntype{C}{>{\centering}p{0.14\textwidth}}
\newcolumntype{D}{>{\centering}p{0.24\textwidth}}
\newcolumntype{E}{>{\centering\arraybackslash}p{0.18\textwidth}}
\begin{tabular}{ABC|DE}
\Xhline{1pt}
Teacher & \makecell{ Pruning \\Ratio} & \makecell{Teacher\\ Accuracy } & Student & \makecell{Student\\Accuracy}\\\Xhline{1pt}
 None         & - & - & VGG19-CL1       & 69.51 $\pm$ 0.24\\\hline
\multirow{4}{*}{VGG19}  & None         & 73.13 & VGG19-CL1    & 70.47 $\pm$ 0.25\\
           & 36\%         & 73.30 & VGG19-CL1     & 71.52 $\pm$ 0.50\\
           & 59\%         & 72.25 & VGG19-CL1     & 71.43 $\pm$ 0.24\\
           & 79\%         & 73.43 & VGG19-CL1     & 71.82 $\pm$ 0.16\\\hline

 None         & - & - & VGG19-CL1    & 69.51 $\pm$ 0.24\\\hline
\multirow{4}{*}{VGG19DBL}  & None        & 74.44  & VGG19-CL1     & 70.38 $\pm$ 0.25\\
  & 36\%       &  73.46  & VGG19-CL1     & 70.84 $\pm$ 0.23\\
  & 59\%       &  73.24  & VGG19-CL1     & 70.52 $\pm$ 0.03\\
  & 79\%       &  73.50  & VGG19-CL1    & 71.00 $\pm$ 0.34\\\hline
  
 None         & - & - & VGG19-CL2      &  71.36 $\pm$ 0.29 \\\hline
\multirow{4}{*}{VGG19}  & None         &  73.13 & VGG19-CL2    & 72.75 $\pm$ 0.60\\
  & 36\%        &  73.30 & VGG19-CL2     &  73.52 $\pm$ 0.22 \\
  & 59\%        &  72.25 & VGG19-CL2    &  73.39 $\pm$ 0.21 \\
  & 79\%        &  73.43 & VGG19-CL2    &  73.67 $\pm$ 0.09 \\\hline
  
 None         & - & - & VGG19-CL2     & 71.36 $\pm$ 0.29\\\hline
\multirow{4}{*}{VGG19DBL}  & None        & 74.44  & VGG19-CL2   & 72.29 $\pm$ 0.12\\
  & 36\%        &  73.46 & VGG19-CL2     & 72.73 $\pm$ 0.41\\
  & 59\%        & 73.24  & VGG19-CL2     & 72.94 $\pm$ 0.37\\
  & 79\%        & 73.50  & VGG19-CL2    & 72.88 $\pm$ 0.20\\
\Xhline{1pt}
\end{tabular}
\end{center}
\end{table*}

Table~\ref{tab:custom-layer} compares the test accuracies of student networks with pruned and unpruned teachers.
The number of parameters of VGG19-CL1 and VGG19-CL2 are 11.0M and 9.9M, respectively,
which are comparable to VGG19-ST59 that has 8.2M parameters (see Appendix for details).
However, the test accuracy of VGG19-ST69 with the proposed framework is
higher than accuracies of VGG19-CL1 and VGG19-CL2.
The result justifies the proposed student network construction based on the pruned teacher.

Also, the student network with pruned teachers outperforms the student with the unpruned teacher.
This implies that the surprising performance of pruned teachers does not rely on the architecture of the student.
Note that VGG19DBL has better test accuracy compared to VGG19, where the margin is about 1\%.
There is no significant difference in test accuracy when unpruned VGG19 and unpruned VGG19DBL are being used as teacher networks in KD.
However, in KD, pruned VGG19 teaches better than pruned VGG19DBL with the same sparsity.
It coincides with what we observed in the previous section, where the teacher with better accuracy does not guarantee better KD.

\subsection{Discussions}
\begin{figure}[t]
\centering
     \begin{subfigure}[b]{0.47\textwidth}
         \centering
         \includegraphics[width=\textwidth]{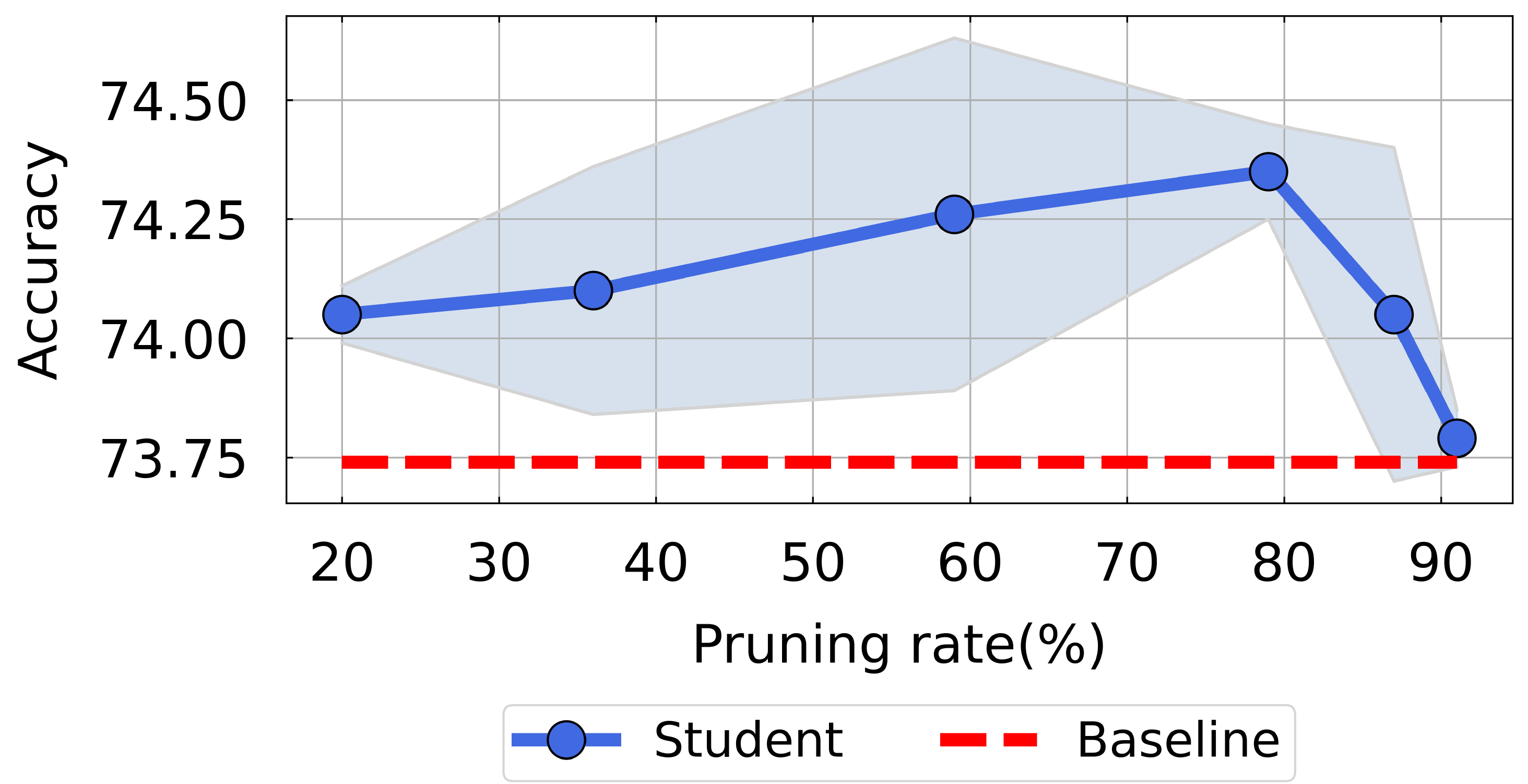}
     \end{subfigure}
     \hfill
     \begin{subfigure}[b]{0.47\textwidth}
         \centering
         \includegraphics[width=\textwidth]{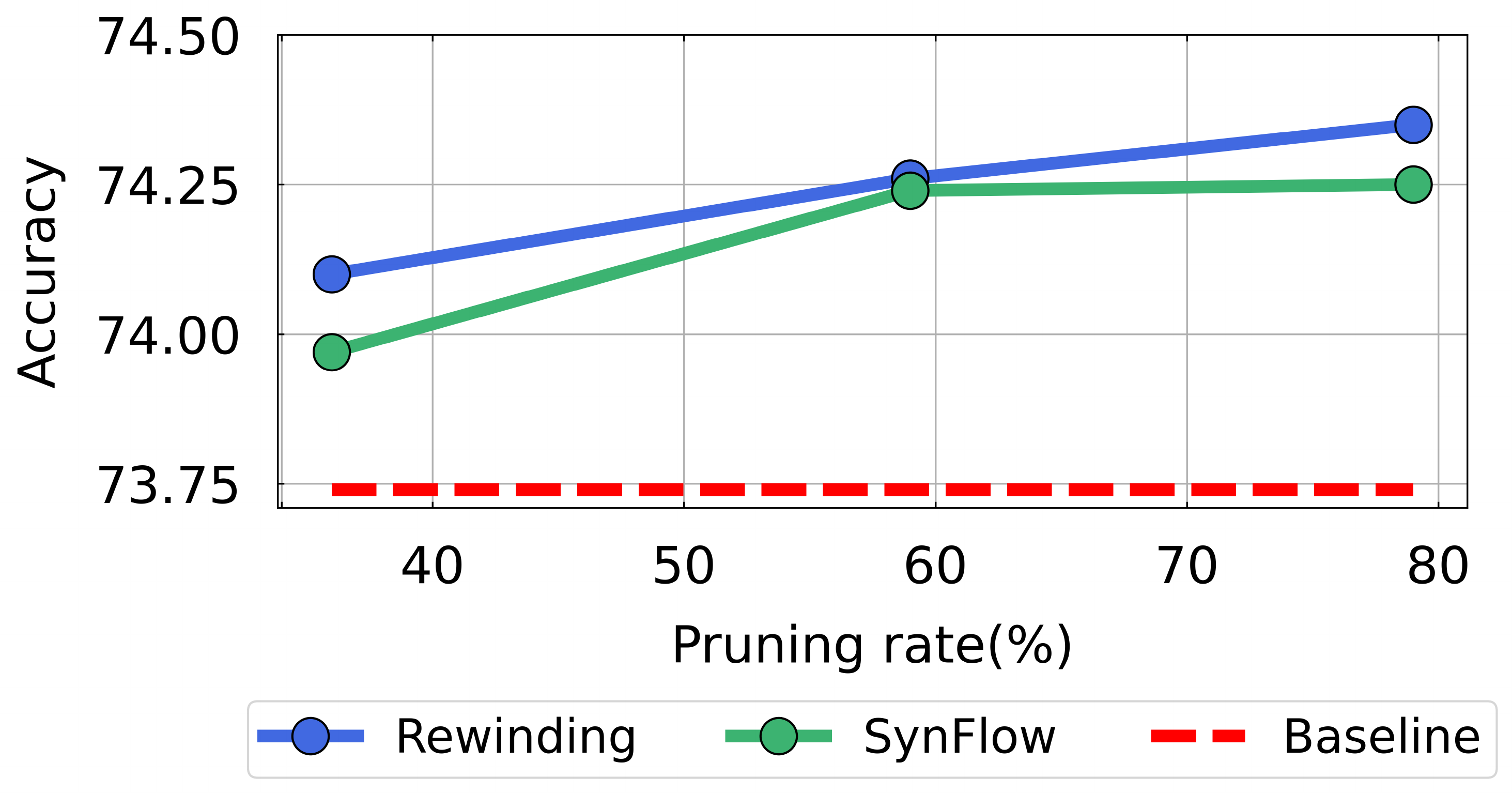}
     \end{subfigure}
     \caption{Effect of pruning ratios and algorithms.
     The left plot shows the student's accuracies with various pruning ratios of pruned teachers.
     The right plot shows the student's accuracies when different pruning algorithms (LR rewinding~\cite{renda2020comparing} and SynFlow~\cite{tanaka2020pruning}) are applied to the teacher.
     In both cases, baseline is the student distilled from unpruned teacher.}
     \label{fig:discussions}
\end{figure}

\noindent{\bf Effect of pruning ratio and pruning algorithm: }
Figure~\ref{fig:discussions} shows the effect of pruning ratio and pruning algorithm.
For VGG19 on CIFAR100, we apply the proposed scheme with additional pruning ratio 20\%, 87\%, and 91\%.
In the current setting, the 79\% point is the optimal pruning ratio,
and the student's performance is degraded if the pruning ratio is too high.
We also applied another pruning algorithm, SynFlow~\cite{tanaka2020pruning}.
Our result shows that the effectiveness of proposed compression scheme does not depend on the choice of pruning algorithm.

\noindent{\bf Large Scale Experiments: }
We also applied the proposed idea to the larger model (ResNet50) and the larger dataset (ImageNet).
We consistently observe that the ``prune, then distill'' strategy is effective in large scale setups as well. 
We refer to the Appendix for a detailed setup and results of large-scale experiments.

\section{Conclusion}
Our experiments showed that the pruned teacher can be more effective than the original teacher in KD.
We further showed theoretically that the pruned teacher provides an additional regularization in distillation.
Based on this observation, we proposed a novel network compression scheme 
that distills a pruned teacher network to the student network whose architecture is based on an (unstructured) pruned network.
The proposed network compression is effectively a structured pruning algorithm that utilizes the knowledge of sparsity from unstructured pruning,
and therefore our work bridges two main pruning approaches.

\section*{Acknowledgments}
JP and AN were supported by Basic Science Research Program through the National Research Foundation of Korea (NRF) funded by the Ministry of Education (2021R1F1A1059567).
We thank Minhyeok Cho for giving valuable comments. We also thank anonymous reviewers for providing constructive feedback.

\clearpage
%
%
\bibliographystyle{main}
\bibliography{main}

\newpage
\appendix

\section{Student Network Design}\label{app:student network design}

Instead of applying optimized network architecture search (NAS),
we use the most naive approach to construct the student network $f_s$.
Given the pruned network $f_t(\cdot;w_p)$ (via unstructured pruning),
we count the number of nonzero parameters for each layer.
Then, the student network $f_s$ is constructed to have the same number of layers as $f_t(\cdot;w_p)$,
but each layer has reduced number of neurons (or channels).
The number of neurons is specifically chosen to (approximately) match the number of parameters per layer
of the pruned network $f_t(\cdot;w_p)$.

Consider the case where the original network is a convolutional neural network (CNN), the most common scenario.
Recall that the number of parameters of a convolutional layer is
\begin{align}
x\times y \times c_{in} \times c_{out}
\end{align}
where $x\times y$ corresponds to the size of the filter, $c_{in}$ is the number of input channels, and $c_{out}$ is the number of output channels.
Note that we ignore the bias for simplicity.
Thus, we can sequentially adjust the number of channels per layer to match the number of parameters.

More precisely, suppose $f_t(\cdot;w_p)$ be a pruned CNN with $L$ layers,
and $n_1, \dots, n_L$ be the number of nonzero parameters in each layer of $f_t(\cdot;w_p)$.
We construct a new student CNN $f_s$ with $L$ layers where the number of channels at each layer is $c_0, c_1, \dots, c_L$
($c_0$ is the number of channels of input, which is 3 for an RGB image).
In each $i$-th layer, the size of filter $x_i\times y_i$ is the same as the pruned CNN $f_P$.
Then, we iteratively match the number of parameters using
\begin{align}
    c_i = \left[\frac{n_i}{x_i\times y_i \times c_{i-1}}\right]
\end{align}
where $[\cdot]$ is a rounding operator.

\clearpage
\section{Training Details}\label{app:training details}
In this section, we describe the detailed experimental setting.
Table~\ref{tab:hparam-train} provide hyperparameters
for regular training, pruning (LR rewinding), and knowledge distillation (vanilla KD), respectively.
Most hyperparameters are common choices in practice.
However, note that we use Nesterov stochastic gradient descent (SGD) as an optimizer
since it is a default optimizer for LR rewinding.
This optimizer may not be an optimal choice, however, our goal is not achieving state-of-the-art test accuracy
but having fair comparison between pruned teacher and unpruned teacher.
For MoblineNetV2, some hyperparameters related to learning rate are modified to ensure accuracy.
Since MobileNetV2 is a student network in our experiment, we do not prune MobileNetV2.

\begin{table*}[h]
\centering
\caption{Hyperparameters for training, pruning, and KD.}
\label{tab:hparam-train}
\begin{tabular}{c||ccc}
\Xhline{1pt}
{\bf Training}      & \multicolumn{1}{c}{VGG} & ResNet & MobileNetV2 \\ \Xhline{1pt}
Optimizer           &  nesterov SGD (0.9)        &   nesterov SGD (0.9) &  nesterov SGD (0.9)                 \\ 
Trainig epochs      &   200                  &  100 &        100           \\
Batch size          &   128                  &  128 &  256              \\
Learning rate       &   0.1                  &  0.01    &  0.05            \\
Learning rate drops &   [60, 120, 160]       &  [30, 60, 80]    & [60, 80]   \\ 
Drop factor         &   0.2                  &  0.1              & 0.1    \\
Weight decay        &   0.0005               &  0.0001            &   0.0005      \\
\Xhline{1pt}
{\bf Pruning}          & \multicolumn{1}{c}{VGG} & ResNet          & - \\ \Xhline{1pt}
Pruner                 &    LR rewinding    &  LR rewinding         & -                 \\ 
Iterative pruning rate &    0.2             &   0.2                 & -  \\ 
Optimizer              &  nesterov SGD (0.9)&   nesterov SGD (0.9)  & -                \\ 
Post trainig epochs    &    130             &    50                 & -  \\ 
Batch size             &    128             &    512                & -   \\ 
Learning rate          &    0.1             & 0.04                  & -\\ 
Learning rate drops    &    [39, 84]        & [10, 30]              & - \\ 
Drop factor            &    0.1             &  0.1                  & -          \\ 
Weight decay           &    0.0002          &     0.0001            & -\\ 
\Xhline{1pt}
{\bf Distillation}& \multicolumn{1}{c}{VGG} & ResNet & MobileNetV2 \\ \Xhline{1pt}
KD                   &  vanilla         &   vanilla    &  vanilla          \\
Optimizer            &  nesterov SGD (0.9)        &   nesterov SGD (0.9)   &   nesterov SGD (0.9)              \\ 
KD epochs            &  200             &   100       &  100         \\
KD batch Size        &  128             &  128      &  256       \\
KD learning Rate     &  0.1             &   0.01 &  0.05      \\
Learning rate drops  &  [60, 120, 160]  &  [30, 60, 80]   & [60, 80] \\
Drop factor          &  0.2             &   0.2   &   0.1 \\
Weight decay         &  0.0005          &   0.0005    &  0.0005            \\
Alpha                &  0.95            &   0.95     &  0.95     \\
Temprature           &  10              &   10       &  10     \\
\Xhline{1pt}
\end{tabular}
\end{table*}

\clearpage

\section{Agreement between Teacher and Student}\label{app:agreement}
In this section, we investigate the agreement between the teacher and student's prediction in various settings.
Table~\ref{tab:agreement} presents the agreement as well as students' accuracy.
As we discussed, increment in agreement does not always guarantee the accuracy.
This implies that the teacher may not ``teach'' the student, but ``help'' the student with regularization.

\begin{table*}
\caption{Agreement between the teacher and the student.}
\label{tab:agreement}
\begin{center}
\newcolumntype{A}{>{\centering}p{0.16\textwidth}}
\newcolumntype{B}{>{\centering}p{0.14\textwidth}}
\newcolumntype{C}{>{\centering}p{0.24\textwidth}}
\newcolumntype{D}{>{\centering}p{0.18\textwidth}}
\newcolumntype{E}{>{\centering\arraybackslash}p{0.18\textwidth}}
\begin{tabular}{AB|CDE}
\Xhline{1pt}
Teacher & \makecell{ Pruning \\Ratio} & Student & \makecell{Student\\Accuracy} & Agreement\\\Xhline{1pt}
\multirow{4}{*}{VGG19}  & None             & VGG19          &  73.74 $\pm$ 0.20 & 76.67 $\pm$ 0.12  \\
                        & 36\%             & VGG19          &  74.10 $\pm$ 0.26 & 77.70 $\pm$ 0.12 \\
                        & 59\%             & VGG19          &  74.26 $\pm$ 0.37 & 77.05 $\pm$ 0.16 \\
                        & 79\%             & VGG19          &  74.35 $\pm$ 0.10 & 78.95 $\pm$ 0.10 \\\hline
\multirow{3}{*}{VGG19}  & 36\%             & VGG19-ST36     &  73.77 $\pm$ 0.16 & 77.09 $\pm$ 0.19 \\
                        & 59\%             & VGG19-ST59     &  73.81 $\pm$ 0.10 & 77.42 $\pm$ 0.42 \\
                        & 79\%             & VGG19-ST79     &  73.39 $\pm$ 0.11 & 77.61 $\pm$ 0.26 \\\hline
\Xhline{1pt}
\multirow{4}{*}{ResNet18}  & None             & ResNet18          &      57.97 $\pm$ 0.10 & 73.91 $\pm$ 0.31\\
                           & 36\%             & ResNet18          &      59.39 $\pm$ 0.21 & 72.07 $\pm$ 0.12\\
                           & 59\%             & ResNet18          &      58.99 $\pm$ 0.26 & 70.79 $\pm$ 0.16\\
                           & 79\%             & ResNet18          &      59.33 $\pm$ 0.18 & 70.57 $\pm$ 0.60 \\\hline
\multirow{3}{*}{ResNet18}  & 36\%             & ResNet18-ST36     &      58.75 $\pm$ 0.19 & 70.59 $\pm$ 0.29 \\
                           & 59\%             & ResNet18-ST59     &      57.76 $\pm$ 0.31 & 68.03 $\pm$ 0.10 \\
                           & 79\%             & ResNet18-ST79     &      56.23 $\pm$ 0.16 & 64.68 $\pm$ 0.27\\\hline
\Xhline{1pt}
\end{tabular}
\end{center}
\end{table*}

\clearpage
\section{Number of Parameters}\label{app:number of parameters}

Table~\ref{tab:model size} shows the number of parameters in networks.
As we intended, we can see that the number of parameters coincides with the target sparsity of pruned teachers.
For example, the number of parameters in VGG19-ST79 is roughly 21\%, matching 79\% sparsity.
We also count FLOPs using ptflops~\cite{ptflops}.
Note that the model with fewer parameters may have more FLOPs.
For example, VGG19-ST79 has fewer weights than VGG19-CL1 but has more FLOPs.
However, VGG19-ST79 shows higher test accuracies, indicating the effectiveness of the student network architecture learned from the pruned teacher.

\begin{table*}[h]
\caption{Number of parameters and FLOPs of various models in our exepriements.}
\label{tab:model size}
\centering
\begin{tabular}{c|c|cc}
\Xhline{1pt}
Datasets & Model     & \# of param & FLOPs \\ \Xhline{1pt}
\multirow{10}{*}{CIFAR100}
&VGG19          &      20.1M & 399M \\ 
&VGG19-CL1      &      11.0M & 158M \\ 
&VGG19-CL2      &       9.9M & 264M\\ 
&VGG19DBL      &      75.4M & 1495M \\ 
&VGG19DBL-ST36  &      48.2M & 1187M\\ 
&VGG19DBL-ST59  &      30.8M & 916M    \\ 
&VGG19DBL-ST79  &      15.7M & 677M    \\ 
&VGG19-ST36     &      12.8M & 321M    \\ 
&VGG19-ST59     &       8.2M & 248M    \\ 
&VGG19-ST79     &       4.2M & 174M    \\ 

\Xhline{1pt}
\multirow{6}{*}{TinyImageNet}
&ResNet18       &    11.3M & 149M    \\ 
&ResNet18-ST36  &     7.3M & 114M \\ 
&ResNet18-ST59  &     4.7M & 91M     \\ 
&ResNet18-ST79  &     2.4M & 66M     \\ 
&VGG16          &    18.1M & 1381M       \\ 
&MobileNetV2   &     2.5M & 27M     \\ 
\Xhline{1pt}
\end{tabular}
\end{table*}

\begin{table*}[h]
\centering
\caption{Number of parameters in each layer of unpruned VGG19, pruned VGG19 (79\%), and VGG19-ST79.}
\label{tab:parameter of each layer}
\begin{tabular}{c||c|cc|cc}
\Xhline{1pt}
\multirow{2}{*}{}& \multicolumn{1}{c}{VGG19}  & \multicolumn{2}{c}{Pruned VGG19 (79\%)} & \multicolumn{2}{c}{VGG19-ST79} \\ 
\hline
        & \# of weight & \# of weight & ratio(\%) & \# of weight & ratio(\%)      \\ \Xhline{1pt}
conv-0  & 1728         & 1087         & 62.91             & 1080         & 62.50    \\
conv-1  & 36864        & 18102        & 49.10             & 17640        & 47.85   \\
conv-2  & 73728        & 50134        & 68.00             & 48951        & 66.39    \\
conv-3  & 147456       & 97936        & 66.42             & 96903        & 65.72     \\
conv-4  & 294912       & 198189       & 67.20             & 196425       & 66.60    \\
conv-5  & 589824       & 381144       & 64.62             & 378675       & 64.20      \\
conv-6  & 589824       & 379358       & 64.32             & 376992       & 63.92     \\
conv-7  & 589824       & 344924       & 58.48             & 342720       & 58.11     \\
conv-8  & 1179648      & 548035       & 46.46             & 544680       & 46.17     \\
conv-9  & 2359296      & 749074       & 31.75             & 746532       & 31.64     \\
conv-10 & 2359296      & 461873       & 19.58             & 461340       & 19.55      \\
conv-11 & 2359296      & 196359       & 8.32              & 196020       & 8.31      \\
conv-12 & 2359296      & 99450        & 4.22              & 98901        & 4.19      \\
conv-13 & 2359296      & 84433        & 3.58              & 83916        & 3.56     \\
conv-14 & 2359296      & 225496       & 9.56              & 224532       & 9.52     \\
conv-15 & 2359296      & 328861       & 13.94             & 326106       & 13.82      \\
fc      & 51200        & 44546        & 87.00             & 12200        & 23.83     \\ \hline{}
total   & 20070088     & 4209001      & 20.97             & 4153613      & 20.70     \\
\Xhline{1pt}
\end{tabular}
\end{table*}

\begin{table*}[h]
\centering
\caption{Number of parameters in each layer of pruned VGG19 (59\%), VGG19-CL1, and VGG19-CL2.}
\label{tab:VGG19-CL1 and VGG19-CL2}
\begin{tabular}{c||c|cc|cc|cc}
\Xhline{1pt}
        & \multicolumn{1}{c}{VGG19} & \multicolumn{2}{c}{Pruned VGG19 (59\%)} & \multicolumn{2}{c}{VGG19-CL1} & \multicolumn{2}{c}{VGG19-CL2} \\
        \hline
        & \# of weight & \# of weight & ratio & \# of weight & ratio & \# of weight & ratio    \\ \Xhline{1pt}
conv-0  & 1728    & 1210         & 70.02 & 1728         & 100   & 1728         & 100      \\
conv-1  & 36864   & 22885        & 62.08 & 36864        & 100   & 22464        & 60.94  \\
conv-2  & 73728   & 59344        & 80.49 & 36864        & 50    & 62829        & 85.22 \\
conv-3  & 147456  & 118013       & 80.03 & 36864        & 25    & 127269       & 86.31 \\
conv-4  & 294912  & 242091       & 82.09 & 73728        & 25    & 251694       & 85.35 \\
conv-5  & 589824  & 487123       & 82.59 & 147456       & 25    & 493830       & 83.72 \\
conv-6  & 589824  & 490757       & 83.2  & 147456       & 25    & 504990       & 85.62 \\
conv-7  & 589824  & 452699       & 76.75 & 147456       & 25    & 475668       & 80.65 \\
conv-8  & 1179648 & 769861       & 65.26 & 294912       & 25    & 806796       & 68.39 \\
conv-9  & 2359296 & 1281396      & 54.31 & 589824       & 25    & 1364922      & 57.85 \\
conv-10 & 2359296 & 1064558      & 45.12 & 589824       & 25    & 1111500      & 47.11 \\
conv-11 & 2359296 & 751546       & 31.85 & 589824       & 25    & 711000       & 30.14 \\
conv-12 & 2359296 & 435158       & 18.44 & 1179648      & 50    & 385362       & 16.33 \\
conv-13 & 2359296 & 380092       & 16.11 & 2359296      & 100   & 339021       & 14.37 \\
conv-14 & 2359296 & 711337       & 30.15 & 2359296      & 100   & 684297       & 29.00 \\
conv-15 & 2359296 & 903232       & 38.28 & 2359296      & 100   & 2520576      & 106.84 \\
fc      & 51200   & 49403        & 96.49 & 51200        & 100   & 51200        & 100      \\ \hline{}
total   & 20070088& 8220705        & 40.96 & 11001536     & 54.82     & 9915146    & 49.40 \\
\Xhline{1pt}
\end{tabular}
\end{table*}


\subsection{VGG-ST}
Table~\ref{tab:parameter of each layer} summarizes the number of weights in each layer for VGG19, pruned VGG19 (79\% sparsity),
and VGG19-ST79.
As we described in SND, We set the number of filters based on the number of weights per layer of the pruned teacher.
Note that we have modified VGG which has a single fully-connected (FC) layer.
We do not control the number of parameters of FC, which is deterministic based on the number of filters in the previous layer.
Thus, the weight ratio of the fc layer does not match the pruned network.
Other student networks, VGG-ST36 and VGG-ST59, were constructed similarly.

\subsection{VGG-CL}
We design VGG19-CL1 and VGG19-CL2 so that the number of parameters of the model is roughly half of the original unpruned model.
For VGG-CL1, we remove half of filters for each layer except conv-0, conv-1, conv-13, conv-14, and conv-15
The role of those layers (that are close to either input or output) are crucial, we keep the whole filters for VGG19-CL1.
VGG-CL1 was designed to check the importance of each layer by remove the channels uniformly across the layers.

For VGG-CL2, we design a network somewhere between pruned VGG19 (59\%) and VGG19-CL1.
Similar to VGG19-CL1, another customized network VGG19-CL2 has the same number of filters in crucial layers (the first and the last).
Thus, conv-15 has 512 filters and an the fully-connected (FC) layer has 51200 weights.
On the other hand, the number of channels in other layers are chosen to match the number of parameters per layer of pruned VGG19 (59\%).
The number of filters for each remaining layer was set to approximate the number of parameters of pruned VGG19 (79\%).
Table~\ref{tab:VGG19-CL1 and VGG19-CL2} shows the number of parameters in each layer of VGG19-CL1 and VGG19-CL2.

\clearpage
\section{Mismatched Pair of Networks}
We apply KD to mixed pair of teacher and student networks.
For example, VGG19-ST36 is a student network that corresponds to pruned VGG19 teacher with sparsity 36\%.
In this section, we transfer knowledge from a teacher to a mismatched student,
for example, the pruned VGG19 teacher with 59\% sparsity when the student is VGG19-ST36.

\begin{table*}[h]
\caption{Distillation between mismatched pair of teacher and student networks (VGG19).}
\label{tab:mixed-cifar}
\begin{center}
\newcolumntype{A}{>{\centering}p{0.20\textwidth}}
\newcolumntype{B}{>{\centering}p{0.14\textwidth}}
\newcolumntype{C}{>{\centering}p{0.14\textwidth}}
\newcolumntype{D}{>{\centering}p{0.24\textwidth}}
\newcolumntype{E}{>{\centering\arraybackslash}p{0.18\textwidth}}
\begin{tabular}{ABC|DE}
\Xhline{1pt}
Teacher & \makecell{ Pruning \\Ratio} & \makecell{Teacher\\ Accuracy } & Student & \makecell{Student\\Accuracy}\\\Xhline{1pt}

None                    & -                            & -                  & VGG19-ST36     &  72.32 $\pm$ 0.12 \\\hline
\multirow{2}{*}{VGG19}  & None                         &   73.13            & VGG19-ST36     & 73.52 $\pm$ 0.20 \\
                        & 36\%                         &   73.30            & VGG19-ST36     & 73.77 $\pm$ 0.16 \\
                        & 59\%                         &   72.25            & VGG19-ST36     & 73.91 $\pm$ 0.15 \\
                        & 79\%                         &   73.43            & VGG19-ST36     & 74.00 $\pm$ 0.20 \\\hline
None                    & -                            & -                  & VGG19-ST59     & 71.80 $\pm$ 0.18 \\\hline
\multirow{2}{*}{VGG19}  & None                         &   73.13            & VGG19-ST59     & 73.18 $\pm$ 0.10 \\
                        & 36\%                         &   73.30            & VGG19-ST59     & 73.42 $\pm$ 0.24 \\
                        & 59\%                         &   72.25            & VGG19-ST59     & 73.81 $\pm$ 0.10  \\
                        & 79\%                         &   73.43            & VGG19-ST59     & 73.69 $\pm$ 0.27 \\\hline
None                    & -                            & -                  & VGG19-ST79     & 70.89 $\pm$ 0.14 \\\hline
\multirow{2}{*}{VGG19}  & None                         &   73.13            & VGG19-ST79     & 72.42 $\pm$ 0.16 \\
                        & 36\%                         &   73.30            & VGG19-ST79     & 72.97 $\pm$ 0.17\\
                        & 59\%                         &   72.25            & VGG19-ST79     & 73.13 $\pm$ 0.09 \\
                        & 79\%                         &   73.43            & VGG19-ST79     & 73.39 $\pm$ 0.11 \\\hline
\Xhline{1pt}
None                       & -                            & -                  & ResNet18-ST36     & 56.44 $\pm$ 0.26 \\\hline
\multirow{2}{*}{ResNet18}  & None                         &  57.75             & ResNet18-ST36     & 57.74 $\pm$ 0.22 \\
                           & 36\%                         &  57.66             & ResNet18-ST36     & 58.75 $\pm$ 0.19 \\
                           & 59\%                         &  57.58             & ResNet18-ST36     & 58.57 $\pm$ 0.22 \\
                           & 79\%                         &  57.32             & ResNet18-ST36     & 58.46 $\pm$ 0.18 \\\hline
None                       & -                            & -                  & ResNet18-ST59     & 55.93 $\pm$ 0.32 \\\hline
\multirow{2}{*}{ResNet18}  & None                         &   57.75            & ResNet18-ST59     & 56.70 $\pm$ 0.35 \\
                           & 36\%                         &   57.66            & ResNet18-ST59     & 58.20 $\pm$ 0.06 \\
                           & 59\%                         &   57.58            & ResNet18-ST59     & 57.76 $\pm$ 0.29\\
                           & 79\%                         &   57.32            & ResNet18-ST59     & 57.94 $\pm$ 0.20 \\\hline
None                       & -                            & -                  & ResNet18-ST79     & 54.48 $\pm$ 0.53 \\\hline
\multirow{2}{*}{ResNet18}  & None                         &  57.75             & ResNet18-ST79     & 55.65 $\pm$ 0.24 \\
                           & 36\%                         &  57.66             & ResNet18-ST79     & 56.66 $\pm$ 0.15\\
                           & 59\%                         &  57.58             & ResNet18-ST79     & 56.19 $\pm$ 0.12\\
                           & 79\%                         &  57.32             & ResNet18-ST79     & 56.23 $\pm$ 0.16 \\\hline
\Xhline{1pt}
\end{tabular}
\end{center}
\end{table*}

Table~\ref{tab:mixed-cifar} shows the result when we mix the teacher and student pair.
Although the student is not designed for the teacher, 
we can see that the pruned teacher teaches better than the unpruned teacher.

\clearpage

\section{Large Scale Experiments}
We conduct a large scale experiment to further justify the proposed algorithm.
Table~\ref{tab:self-distillation50} shows the self distillation result of ResNet50 with and without pruned teacher.
It is clear that distillation from pruned teacher is better than the distillation from unpruned teacher.

\begin{table*}[h]
\begin{center}
\caption{Self distillation of ResNet50 with teacher pruning. 
Teacher ``None'' indicates the student is trained without a teacher,
while the pruning ratio ``None'' means the distillation from the unpruned teacher.
}
\label{tab:self-distillation50}
\newcolumntype{A}{>{\centering}p{0.20\textwidth}}
\newcolumntype{B}{>{\centering}p{0.16\textwidth}}
\newcolumntype{C}{>{\centering}p{0.16\textwidth}}
\newcolumntype{D}{>{\centering}p{0.20\textwidth}}
\newcolumntype{E}{>{\centering\arraybackslash}p{0.18\textwidth}}
\begin{tabular}{ABC|DE}
\Xhline{1pt}
Teacher & \makecell{ Pruning \\Ratio} & \makecell{Teacher\\ Accuracy } & Student & \makecell{Student\\Accuracy}\\\Xhline{1pt}
  
None & - & - & ResNet50    & 62.88 $\pm$ 0.25 \\\hline
\multirow{4}{*}{ResNet50}  & None      & 62.88 & ResNet50     & 64.54 $\pm$ 0.35\\
  & 36\%         & 62.72  & ResNet50     & 64.86 $\pm$ 0.06\\
  & 59\%        &  62.85 & ResNet50     & 65.21 $\pm$ 0.21\\
  & 79\%       &  63.46    & ResNet50     & 64.97 $\pm$ 0.10\\

\Xhline{1pt}
\end{tabular}
\end{center}
\end{table*}

Table~\ref{tab:algorithm-resnet50} shows the performance of the proposed compression algorithm on ResNet50 with TinyImageNet.
Similar to our main experiment with ResNet18 models, we also observe the effectiveness of our scheme in the larger model.
\begin{table*}[h]
\caption{Performance of the proposed compression algorithm on ResNet50 with TinyImageNet.
ResNet50-ST(X) is the constructed student network based on the proposed algorithm from X\% pruned teacher.
Teacher ``None'' indicates the student is trained without a teacher,
while the pruning ratio ``None'' means the distillation from the unpruned teacher.
}
\label{tab:algorithm-resnet50}
\begin{center}
\newcolumntype{A}{>{\centering}p{0.20\textwidth}}
\newcolumntype{B}{>{\centering}p{0.12\textwidth}}
\newcolumntype{C}{>{\centering}p{0.12\textwidth}}
\newcolumntype{D}{>{\centering}p{0.28\textwidth}}
\newcolumntype{E}{>{\centering\arraybackslash}p{0.18\textwidth}}
\begin{tabular}{ABC|DE}
\Xhline{1pt}
Teacher & \makecell{ Pruning \\Ratio} & \makecell{Teacher\\ Accuracy } & Student & \makecell{Student\\Accuracy}\\\Xhline{1pt}
None & - & - & ResNet50-ST36     & 62.24 $\pm$ 0.14\\\hline
\multirow{2}{*}{ResNet50}  & None         & 62.88 & ResNet50-ST36    & 64.11 $\pm$ 0.26\\
           & 36\%        &  62.72 & ResNet50-ST36   & 64.12  $\pm$ 0.30\\\hline

None & - & - & ResNet50-ST59   & 60.04 $\pm$ 0.29\\\hline
\multirow{2}{*}{ResNet50}  & None         & 62.88 & ResNet50-ST59   & 63.84 $\pm$ 0.17\\
           & 59\%         & 62.85 & ResNet50-ST59 &  63.58 $\pm$ 0.29\\\hline

None & - & - & ResNet50-ST79       & 58.74 $\pm$ 0.06\\\hline
\multirow{2}{*}{ResNet50}  & None          & 62.88 & ResNet50-ST79  & 62.25 $\pm$ 0.46\\
           & 79\%          & 63.46 & ResNet50-ST79  & 62.84 $\pm$ 0.26\\

\Xhline{1pt}
\end{tabular}
\end{center}
\end{table*}

We also run an experiment with ImageNet, which is larger and realistic dataset.
Table~\ref{tab:algorithm-imagenet} shows the performance of the proposed compression algorithm on ResNet18 with ImageNet.
For the pruning ratio of 36\%, the pruned teacher performs better than the unpruned teacher as we observed in the previous experiments.
However, for the pruning ratio of 79\%, the pruned teacher is not effective, mainly because ResNet18 is not sufficiently large for the ImageNet dataset.
This result emphasizes the importance of finding the right pruning ratio for the teacher.

\begin{table*}[h]
\caption{Performance of the proposed compression algorithm on ResNet18 with ImageNet.
The pruning ratio ``None'' means the distillation from the unpruned teacher.
}
\label{tab:algorithm-imagenet}
\begin{center}
\newcolumntype{A}{>{\centering}p{0.20\textwidth}}
\newcolumntype{B}{>{\centering}p{0.12\textwidth}}
\newcolumntype{C}{>{\centering}p{0.12\textwidth}}
\newcolumntype{D}{>{\centering}p{0.28\textwidth}}
\newcolumntype{E}{>{\centering\arraybackslash}p{0.18\textwidth}}
\begin{tabular}{ABC|DE}
\Xhline{1pt}

Teacher & \makecell{ Pruning \\Ratio} & \makecell{Teacher\\ Accuracy } & Student & \makecell{Student\\Accuracy}\\\Xhline{1pt}
\multirow{2}{*}{ResNet18}  & None         & 64.90 & ResNet18-ST36    & 60.93\\
           & 36\%        &  65.41 & ResNet18-ST36   & 61.10\\\hline
\multirow{2}{*}{ResNet18}  & None          & 64.90 & ResNet18-ST79  & 50.24\\
           & 79\%          & 64.70 & ResNet18-ST79  & 50.14\\

\Xhline{1pt}
\end{tabular}
\end{center}
\end{table*}

\end{document}